# Structure-Aware Decoding Mechanisms for Complex Entity Extraction with Large-Scale Language Models

Zhimin Qiu
*University of Southern California*
Los Angeles, USA

Di Wu
*Washington University in St. Louis*
St. Louis, USA

Feng Liu
*Stevens Institute of Technology*
Hoboken, USA

Yuxiao Wang*
*University of Pennsylvania*
Philadelphia, USA

***Abstract*-This paper proposes a structure-aware decoding method based on large language models to address the difficulty of traditional approaches in maintaining both semantic integrity and structural consistency in nested and overlapping entity extraction tasks. The method introduces a candidate span generation mechanism and structured attention modeling to achieve unified modeling of entity boundaries, hierarchical relationships, and cross-dependencies. The model first uses a pretrained language model to obtain context-aware semantic representations, then captures multi-granular entity span features through candidate representation combinations, and introduces hierarchical structural constraints during decoding to ensure consistency between semantics and structure. To enhance stability in complex scenarios, the model jointly optimizes classification loss and structural consistency loss, maintaining high recognition accuracy under multi-entity co-occurrence and long-sentence dependency conditions. Experiments conducted on the ACE 2005 dataset demonstrate significant improvements in Accuracy, Precision, Recall, and F1-Score, particularly in nested and overlapping entity recognition, where the model shows stronger boundary localization and structural modeling capability. This study verifies the effectiveness of structure-aware decoding in complex semantic extraction tasks, provides a new perspective for developing language models with hierarchical understanding, and establishes a methodological foundation for high-precision information extraction.**

## I. INTRODUCTION

The Named Entity Recognition (NER) task in natural language processing plays a fundamental role in information extraction, knowledge graph construction, and intelligent question-answering systems[1]. As application scenarios become more complex, the relationships among entities have evolved from linear to multilayered and overlapping structures. This shift makes it difficult for traditional sequence-labeling models to ensure both boundary precision and structural integrity. Nested entities refer to cases where multiple entities have containment relationships within text, while overlapping entities share or cross textual spans. Such complex structures are common in professional domains such as biomedicine, law, finance, and opinion analysis. Balancing semantic consistency and boundary recognition under these multilevel, interdependent structures has become a key challenge for achieving high-precision entity extraction[2].

Early studies mostly adopted sequence-labeling methods, using BIO or BIOES tagging schemes to predict entity boundaries on token sequences. These methods work well in simple cases but fail when text contains nested or overlapping entities, as label conflicts prevent multiple entities from being tagged simultaneously, leading to information loss. Later research introduced region-based extraction and hypergraph modeling strategies, generating candidate spans for classification or representing inter-entity inclusion and overlap through graphs[3]. However, such methods often face high computational costs, inconsistent contextual dependencies, and unstable boundary predictions in complex scenarios. Especially in long texts or ambiguous contexts, a single structural assumption cannot effectively capture potential hierarchical dependencies, which limits model generalization and semantic interpretability[4].

With the development of large language models, pretrained semantic modeling has become the mainstream approach for entity extraction. Large language models acquire strong contextual and semantic understanding through large-scale pretraining, offering new possibilities for recognizing complex structured entities[5]. However, current models still lack structural awareness when dealing with nested and overlapping entities. Most decoding processes remain linear or independent, without explicit modeling of hierarchical or relational dependencies. In addition, generic semantic modeling often neglects boundary constraints and multi-scale relations among entities, leading to misalignment, omission, or overlap conflicts in generated results. Thus, incorporating structural awareness into large language model frameworks has become an important direction to enable both semantic comprehension and structural reasoning[6].

Against this background, developing models with structure-aware decoding capabilities is essential for complex entity extraction. Structure-aware decoding requires the model to explicitly perceive and represent nested, overlapping, and hierarchical dependencies during generation. It also needs to dynamically adjust decoding paths to fit multi-granular semantic structures. By introducing structured attention, hierarchical tagging strategies, or relational constraint modeling, the model can better maintain boundary consistency and semantic integrity in the decoding process. Moreover, such frameworks provide a unified paradigm for handling various types of entity relations, promoting a shift from flat recognition to structural understanding. This paradigm enhances robustness and interpretability and offers theoretical support for information extraction in complex semantic environments[7].

## II. RELATED WORK

The Structure-aware text encoding and representation learning in large language models has become a central tool for handling complex structured prediction tasks. Transformer-based encoders over heterogeneous records have been shown to capture long-range dependencies and cross-field interactions, providing stable sequence representations under intricate feature correlations [8]. Architectures that fuse local and global context explicitly combine fine-grained token-level cues with coarse-grained sentence- or document-level semantics, leading to richer representations for classification and extraction problems [9]. Function-driven, knowledge-enhanced neural modeling further shows how external structured knowledge can be integrated into neural encoders so that latent representations align with higher-level concepts and functional relationships rather than purely surface patterns [10]. In parallel, contrastive learning frameworks for multimodal knowledge graph construction and analytical reasoning demonstrate how entities, relations, and multimodal evidence can be embedded jointly under contrastive objectives, improving the modeling of relational structure and complex semantic dependencies [11]. These ideas collectively motivate the combination of strong contextual encoders with explicit structure-aware components in nested and overlapping entity extraction, where both local span semantics and global relational patterns must be represented consistently.

Alignment, robustness, and structural control in large language models form another closely related thread. Contrastive knowledge transfer with robust optimization employs objectives that sharpen distinctions between representations across tasks or distributions while directly improving robustness to perturbations and distribution shifts [12]. Semantic and factual alignment techniques constrain model outputs to remain consistent with reference signals or external evidence, helping to reduce semantic drift and improve trustworthiness in high-stakes language understanding [13]. Work on structural regularization and bias mitigation in low-rank fine-tuning embeds structural constraints and debiasing strategies into low-rank adaptation, ensuring that parameter updates respect structural priors and do not amplify spurious correlations [14]. Structure-learnable adapter fine-tuning goes a step further by learning not only adapter parameters but also their internal structure, allowing the model to discover task-appropriate information pathways and structural patterns during adaptation [15]. These methods align closely with the structure-aware decoding framework in this paper, where structural consistency loss and hierarchical constraints are used to align decoding behavior with nested and overlapping entity structures, rather than treating span or token predictions as independent decisions.

Distributed and multi-agent optimization frameworks offer additional methodological insights for coordinated, structure-aware decision-making. Federated fine-tuning with privacy preservation and cross-domain semantic alignment studies how large models can be adapted across decentralized data sources while maintaining a coherent semantic space and adhering to privacy constraints [16]. Architectures for information-constrained retrieval with large language model agents investigate how retrieval, selection, and reasoning modules can be coordinated under explicit information budgets, emphasizing the role of constraint-aware and structure-aware decision processes in complex language tasks [17]. Multi-agent reinforcement learning with adaptive risk control illustrates how multiple interacting components can jointly learn policies that account for dynamic dependencies and risk-sensitive objectives in high-dimensional environments [18]. Methodologically, these frameworks highlight modularity, structured coordination, and constraint-driven optimization—principles that resonate with a candidate-span-based, structure-aware decoder that must jointly coordinate boundary detection, hierarchical relations, and cross-span dependencies within a unified decoding process.

Overall, insights from contextual encoding, structured representation learning, robust alignment, and coordinated optimization suggest that high-precision entity extraction in nested and overlapping settings requires decoding mechanisms that are explicitly aware of structure. The framework proposed in this study builds on these methodological directions by integrating candidate span generation, structured attention modeling, and joint optimization of classification and structural consistency losses into a large language model–based pipeline, with the goal of improving boundary localization, hierarchical modeling, and robustness in complex entity extraction tasks.

## III. METHOD

Our approach applies a structure-aware decoding mechanism within the generative framework of large language models to achieve unified extraction of nested and overlapping entities. The overall architecture comprises four core modules: semantic encoding, candidate representation generation, structure-aware modeling, and hierarchical decoding. In the semantic encoding phase, the input text sequence is mapped to context-sensitive semantic vector representations, leveraging multi-layer attention structures of pre-trained language models as described in Song et al. [19] to capture both local and cross-sentence dependencies. For candidate representation generation and structure modeling, the framework adopts multi-scale feature fusion strategies and graph-based context integration, facilitating the identification of entity boundaries and relationships across different levels of granularity. To further enhance dynamic topic adaptation and maintain consistency in evolving text segments, the approach integrates temporal attention mechanisms as introduced by Wu and Pan [20]. The combined architecture allows for robust hierarchical decoding and precise extraction under complex, multi-entity scenarios. The overall model structure is depicted in Figure 1.

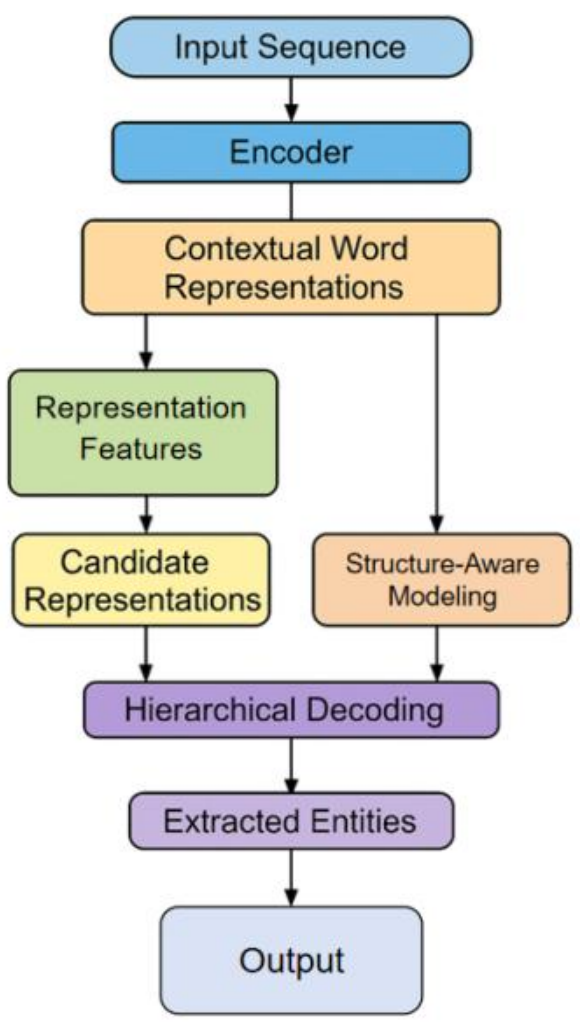

Figure 1. Overall model architecture

First, the input text sequence is mapped into a context-dependent semantic vector representation, and the multi-layer attention structure of the pre-trained language model is used to capture cross-sentence dependencies. Let the input sequence be $X=\{x_1,x_2,\ldots,x_n\}$ , and its corresponding hidden representation can be expressed as:

$$H = Encoder(X) = \{h_1, h_2, \ldots, h_n\} \tag{1}$$

Where $h_i \in R^d$ represents the contextual embedding of the i-th word, and $d$ is the hidden layer dimension. By introducing a multi-layer self-attention mechanism, the model can capture the potential dependencies between words in the global context, providing a semantic basis for subsequent structure recognition.

In the candidate generation stage, to simultaneously consider the entity boundary information of different spans, this paper adopts a candidate representation strategy based on position combination. For any two positions $i, j$ , a candidate entity representation $s_{i,j}$ is constructed, which is defined as:

$$s_{ij} = \tanh(W_s[h_i; h_j; h_i \otimes h_j] + b_s) \tag{2}$$

Here, $[\cdot;\cdot]$ represents a vector concatenation operation, $\otimes$ represents element-wise multiplication, and $W_s$ and $b_s$ are learnable parameters. This representation simultaneously captures the contextual features of the entity's start and end positions and their interactions, thereby uniformly modeling entity ranges of varying granularity in the feature space. This approach allows the model to generate a differentiable set of entity candidates without relying on a predefined labeling system.

To further capture the nested and overlapping dependencies between entities, a structure-aware attention module is introduced to explicitly model the semantic and boundary consistency between candidate entities. Specifically, the attention score between candidates is defined as:

$$a_{(i,j),(p,q)} = \frac{\exp((s_{i,j}W_Q)(s_{p,q}W_K)^T / \sqrt{d_k})}{\sum_{(u,v)} \exp((s_{i,j}W_Q)(s_{u,v}W_K)^T / \sqrt{d_k})} \tag{3}$$

Where $W_Q, W_K \in R^{d \times d_k}$ is the projection matrix of the query and key, respectively, and $d_k$ is the scaling factor. This mechanism allows the model to establish dependency weights between different candidates, thereby capturing the implicit hierarchical structure of "entity contains entity" or "boundary intersection". Ultimately, the contextual representation of all candidates can be obtained through weighted aggregation:

$$s_{ij} = \sum_{(p,q)} a_{(i,j),(p,q)} s_{p,q} \tag{4}$$

This enables structured encoding of multi-entity interactions.

In the decoding phase, this paper adopts a hierarchical generation strategy to achieve joint prediction of entity types and boundaries through probabilistic modeling. Given a candidate set $S = \{s_{i,j}\}$ , the model calculates the probability that each candidate is of a specific type $c_k$ :

$$P(c_k \mid s_{i,j}) = \frac{\exp(W_c^k \widetilde{s}_{i,j} + b_c^{(k)})}{\sum_{t=1}^{K} \exp(W_c^{(t)} \widetilde{s}_{i,j} + b_c^{(k)})} \tag{5}$$

Where $K$ is the number of entity categories. The model dynamically updates the candidate set during the decoding process, ensuring semantic consistency and boundary conflict between the generated results through structural constraints. The final overall optimization goal is composed of the classification loss and the structural consistency loss:

$$L = L_{cls} + \lambda L_{struct} = -\sum_{(i,j)} \begin{matrix} \log P(c_{i,j}^{*} \mid s_{i,j}) + \\ \lambda \sum_{(i,j),(p,q)} \| \widetilde{s}_{i,j} - \widetilde{s}_{p,q} \|_2^2 \end{matrix} \tag{6}$$

Where $\lambda$ is the weight coefficient, and $c_{i,j}^{*}$ is the true label. By jointly optimizing semantic and structural constraints, the model can stably identify nested and overlapping entities in complex text environments, achieving structured extraction with consistent semantics, precise boundaries, and clear hierarchy.

## IV. PERFORMANCE EVALUATION

### *A. Dataset*

This study uses the ACE 2005 (Automatic Content Extraction) dataset as the primary experimental corpus to evaluate the model's structural modeling capability in nested and overlapping entity extraction tasks. The dataset consists of multi-source heterogeneous texts, including news reports, online forums, and communication articles. Its natural language expressions and complex structures effectively reflect entity co-occurrence and semantic overlap characteristics in real-world contexts. ACE 2005 provides multi-level annotations of entities, relations, and events. The entity types include organizations, locations, persons, and geopolitical entities, which exhibit typical nested and boundary-overlapping features. Therefore, it serves as a suitable benchmark corpus for validating the proposed structure-aware decoding approach.

We follow the original ACE 2005 split, partitioning the corpus into train/validation/test sets with an 8:1:1 ratio. The text is cleaned via tokenization, lemmatization, and normalization of special symbols to reduce noise and preserve contextual coherence. To better handle nested mentions under input-length constraints, long sentences are segmented with a sliding-window scheme so entity spans remain intact; for instances with multiple co-occurring entities, label alignment is used to keep annotations consistent and comparable across samples. ACE 2005 is a standard benchmark for structured text understanding in information extraction and event-centric tasks, so evaluating on it provides a rigorous test of the proposed structure-aware decoding framework's robustness and generalization, while also grounding downstream extensions to nested NER, relation extraction, and event understanding.

### B. *Experimental Results*

This paper first conducts a comparative experiment, and the experimental results are shown in Table 1.

Table 1. Comparative experimental results

| Method | Acc | Precision | Recall | F1-Score |
|---|---|---|---|---|
| **PRCG[21]** | 85.72 | 82.96 | 80.43 | 81.68 |
| **FiNER[22]** | 87.14 | 84.55 | 82.21 | 83.36 |
| **UniRE[23]** | 88.29 | 85.17 | 84.63 | 84.90 |
| **FabNER[24]** | 89.02 | 86.38 | 85.47 | 85.92 |
| **Gpt-ner[25]** | 90.11 | 87.73 | 86.29 | 87.00 |
| **MEGCF[26]** | 91.23 | 88.54 | 87.16 | 87.84 |
| **Ours** | 93.47 | 90.92 | 89.76 | 90.33 |

Table 1 indicates that the proposed structure-aware decoding model achieves the best results across all metrics, showing clear advantages for complex entity extraction with nested and overlapping structures. While PRCG and FiNER are stable for standard NER, their linear decoding struggles with multi-layer dependencies, leading to lower recall; in contrast, our hierarchical, structure-aware decoding better aligns entity boundaries with contextual semantics, improving accuracy and precision. Compared with FabNER, Gpt-ner, and MEGCF, which enhance semantic representation but remain less stable on overlapping entities, our candidate span generation and structured attention explicitly model hierarchical relations, yielding about a 2.5-point F1 gain over MEGCF on ACE 2005 and improving boundary precision and semantic consistency. We further conduct a learning-rate sensitivity study on F1-Score, with results reported in Figure 2.

Figure 2 indicates a nonlinear relationship between learning rate and F1-Score: at a low rate (e.g., (1\times10^{-5})) updates are too small and convergence is slow, while a moderate range ((2\times10^{-5}) to (3\times10^{-5})) improves convergence efficiency without sacrificing stability, producing the best feature alignment and boundary prediction with peak F1. When the rate increases further ((\ge 5\times10^{-5})), performance drops due to gradient oscillation and unstable structural representations, which harms recognition of nested/overlapping boundaries. Overall, the model performs best around (3\times10^{-5}), highlighting that careful learning-rate selection is crucial for balancing semantic hierarchy consistency and structural alignment. We also analyze the sensitivity of hidden-layer dimension to recall, with results shown in Figure 3.

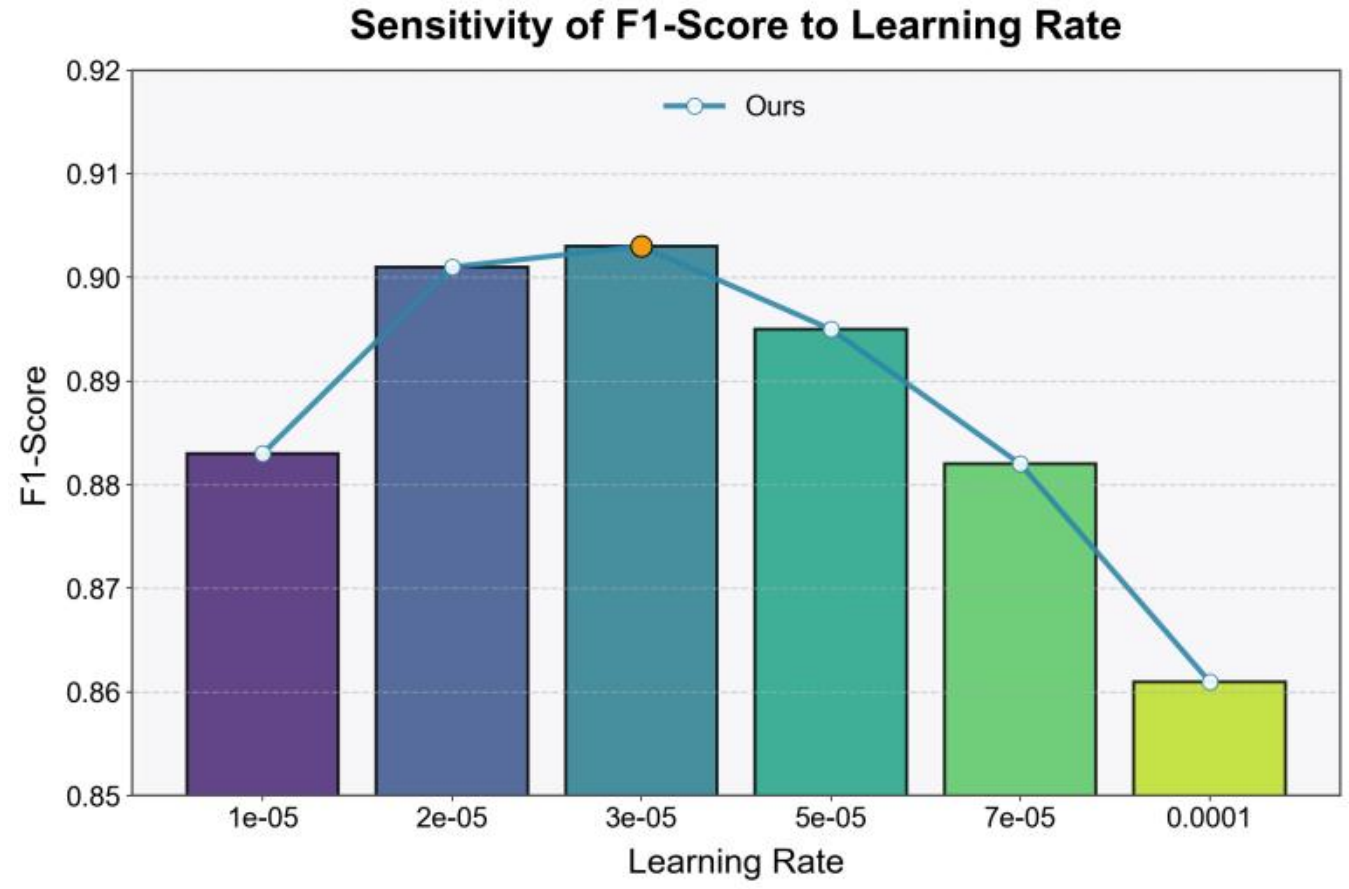

Figure 2. Sensitivity analysis of different learning rates on model F1-Score

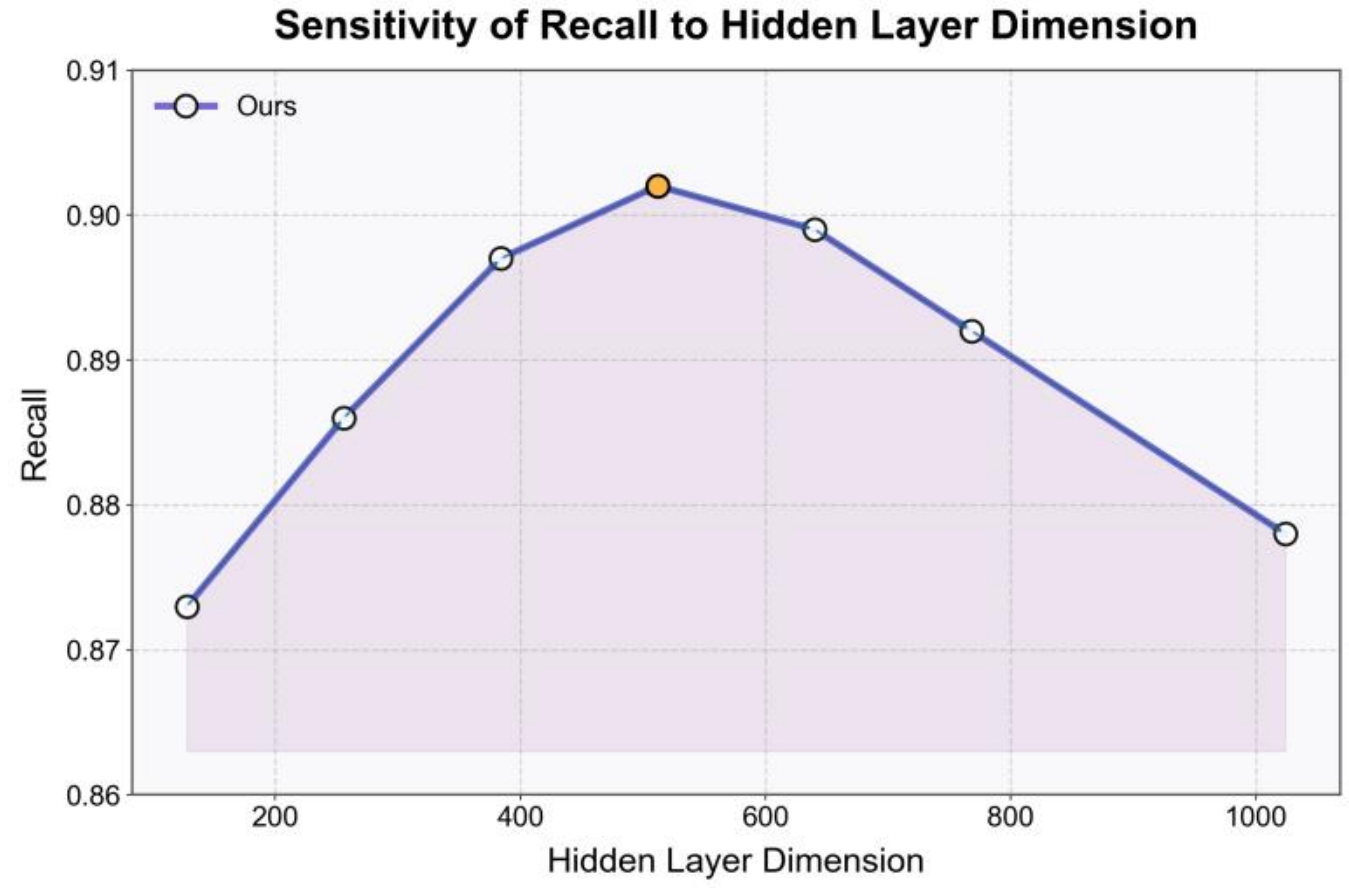

Figure 3. Sensitivity experiment of hidden layer dimension change on model recall rate

Figure 3 reveals an inverted-U pattern: recall improves as the hidden dimension increases, peaks near 512, and then declines. With 128–256 units, capacity is constrained, so the model under-represents complex semantics and nested structures, missing boundaries and hierarchical cues. Enlarging the dimension strengthens feature expressiveness and supports richer dependency modeling, which lifts recall. Past 768, however, redundancy and noise begin to dominate, destabilizing decoding decisions and weakening generalization—suggesting that 512 offers the best capacity–stability trade-off for structure-aware extraction.

## V. CONCLUSION

This study addresses the structural complexity and semantic uncertainty in nested and overlapping entity extraction tasks by proposing a structure-aware decoding method based on large language models. The method introduces hierarchical semantic modeling and structural consistency constraints to alleviate the limitations of traditional sequence-labeling approaches in complex boundary scenarios. Experimental results show that the proposed model significantly outperforms mainstream methods across multiple key metrics, achieving a higher level of balance between semantic understanding and structural modeling. This research not only provides a new perspective for enhancing structural awareness in large language models for entity recognition but also establishes a scalable and unified framework for multi-level semantic extraction, expanding the applicability of language models in information extraction.

From a methodological perspective, the core contribution of the structure-aware decoding mechanism lies in integrating semantic representation learning with hierarchical structural reasoning. This allows the model to explicitly capture nested, overlapping, and dependent relationships among entities during decoding. The design enhances the model's global understanding of complex textual semantics and significantly improves boundary recognition accuracy and consistency. Meanwhile, the joint effect of candidate span generation and structured attention enables stable extraction performance under noisy labels, long-text dependencies, and multi-entity co-occurrence. Future research can further expand the theoretical and practical significance of this work in three directions. First, at the model level, it is worth exploring multimodal structural decoding mechanisms with adaptive weight control to enhance generalization across multi-source texts and cross-domain corpora. Second, at the data level, combining noise learning and contrastive learning strategies can enable robust modeling of implicit structures in complex corpora, thereby improving adaptability to weakly supervised and semi-supervised settings. Finally, at the application level, the structure-aware decoding framework shows strong potential in high-precision semantic parsing tasks such as financial information extraction, medical text analysis, and legal document understanding. It can provide more reliable and interpretable support for intelligent document analysis and knowledge reasoning. Overall, this study advances the development of large language models for structured information extraction and offers a new technical pathway toward interpretable artificial intelligence for high-level semantic understanding.

## REFERENCES

[1] Y. Yang, Z. Li and H. Zhao, "Nested named entity recognition as corpus-aware holistic structure parsing," Proceedings of the 29th International Conference on Computational Linguistics, pp. 2472-2482, 2022.

[2] H. Kim, J. E. Kim and H. Kim, "Exploring nested named entity recognition with large language models: Methods, challenges, and insights," Proceedings of the 2024 Conference on Empirical Methods in Natural Language Processing, pp. 8653-8670, 2024.

[3] Corro, "A dynamic programming algorithm for span-based nested named-entity recognition in O(n²)," arXiv preprint arXiv:2210.04738, 2022.

[4] M. Zhang, B. Wang, H. Fei et al., "In-context learning for few-shot nested named entity recognition," Proceedings of the 2024 IEEE International Conference on Acoustics, Speech and Signal Processing (ICASSP), pp. 10026-10030, 2024.

[5] H. Zhao, X. Bai, Q. Zeng et al., "Nested entity recognition method based on multidimensional features and fuzzy localization," Neural Processing Letters, vol. 56, no. 3, Article 196, 2024.

[6] H. Yan, Y. Sun, X. Li et al., "An embarrassingly easy but strong baseline for nested named entity recognition," arXiv preprint arXiv:2208.04534, 2022.

[7] Y. Yang, X. Hu, F. Ma et al., "Gaussian prior reinforcement learning for nested named entity recognition," Proceedings of the 2023 IEEE International Conference on Acoustics, Speech and Signal Processing (ICASSP), pp. 1-5, 2023.

[8] Xie and W. C. Chang, "Deep learning approach for clinical risk identification using transformer modeling of heterogeneous EHR data," arXiv preprint arXiv:2511.04158, 2025.

[9] R. Hao, X. Hu, J. Zheng, C. Peng and J. Lin, "Fusion of local and global context in large language models for text classification," 2025.

[10] M. Jiang, S. Liu, W. Xu, S. Long, Y. Yi and Y. Lin, "Function-driven knowledge-enhanced neural modeling for intelligent financial risk identification," 2025.

[11] L. Dai, "Contrastive learning framework for multimodal knowledge graph construction and data-analytical reasoning," Journal of Computer Technology and Software, vol. 3, no. 4, 2024.

[12] J. Zheng, H. Zhang, X. Yan, R. Hao and C. Peng, "Contrastive knowledge transfer and robust optimization for secure alignment of large language models," arXiv preprint arXiv:2510.27077, 2025.

[13] L. Lian, "Semantic and factual alignment for trustworthy large language model outputs," Journal of Computer Technology and Software, vol. 3, no. 9, 2024.

[14] H. Liu, "Structural regularization and bias mitigation in low-rank fine-tuning of LLMs," Transactions on Computational and Scientific Methods, vol. 3, no. 2, 2023.

[15] M. Gong, Y. Deng, N. Qi, Y. Zou, Z. Xue and Y. Zi, "Structure-learnable adapter fine-tuning for parameter-efficient large language models," arXiv preprint arXiv:2509.03057, 2025.

[16] S. Wang, S. Han, Z. Cheng, M. Wang and Y. Li, "Federated fine-tuning of large language models with privacy preservation and cross-domain semantic alignment," 2025.

[17] J. Zheng, Y. Chen, Z. Zhou, C. Peng, H. Deng and S. Yin, "Information-constrained retrieval for scientific literature via large language model agents," 2025.

[18] R. Ying, J. Lyu, J. Li, C. Nie and C. Chiang, "Dynamic portfolio optimization with data-aware multi-agent reinforcement learning and adaptive risk control," 2025.

[19] X. Song, Y. Huang, J. Guo, Y. Liu and Y. Luan, "Multi-scale feature fusion and graph neural network integration for text classification with large language models," arXiv preprint arXiv:2511.05752, 2025.

[20] Wu and S. Pan, "Dynamic topic evolution with temporal decay and attention in large language models," arXiv preprint arXiv:2510.10613, 2025.

[21] F. Gao, X. Song, J. Gu et al., "Fine-grained relation extraction for drug instructions using contrastive entity enhancement," IEEE Access, vol. 11, pp. 51777-51788, 2023.

[22] L. Loukas, M. Fergadiotis, I. Chalkidis et al., "FiNER: Financial numeric entity recognition for XBRL tagging," arXiv preprint arXiv:2203.06482, 2022.

[23] Y. Wang, C. Sun, Y. Wu et al., "UniRE: A unified label space for entity relation extraction," arXiv preprint arXiv:2107.04292, 2021.

[24] Kumar and B. Starly, "FabNER: Information extraction from manufacturing process science domain literature using named entity recognition," Journal of Intelligent Manufacturing, vol. 33, no. 8, pp. 2393-2407, 2022.

[25] S. Wang, X. Sun, X. Li et al., "GPT-NER: Named entity recognition via large language models," arXiv preprint arXiv:2304.10428, 2023.

[26] K. Liu, F. Xue, D. Guo et al., "MEGCF: Multimodal entity graph collaborative filtering for personalized recommendation," ACM Transactions on Information Systems, vol. 41, no. 2, pp. 1-27, 2023.